\documentclass[10pt,twocolumn,letterpaper]{article}

\usepackage{iccv}
\usepackage{times}
\usepackage{epsfig}
\usepackage{graphicx}
\usepackage{amsmath}
\usepackage{amssymb}
\usepackage{caption}
\usepackage{color}
\usepackage{comment}
\usepackage{enumitem}
\usepackage{multirow}

\def\bfb{{\bf b}} 
\def\GraviCap{\textsc{GraviCap}}
\newcommand{\norm}[1]{\left\lVert#1\right\rVert}
\usepackage[pagebackref=true,breaklinks=true,letterpaper=true,colorlinks,bookmarks=false]{hyperref}
\usepackage[labelsep=period, labelfont=bf, font=small]{caption} 
\iccvfinalcopy

\begin{document}

\title{Gravity-Aware Monocular 3D Human-Object Reconstruction} 

\author{
Rishabh Dabral$^{1,2}$ \hspace{0.95em}
Soshi Shimada$^2$ \hspace{0.85em} 
Arjun Jain$^{3,4}$ \hspace{0.65em}
Christian Theobalt$^2$ \hspace{0.75em}
Vladislav Golyanik$^2$ \vspace{8pt}\\
$^1$IIT Bombay \hspace{2.1em} 
$^2$MPI for Informatics, SIC \hspace{2.1em}
$^3$IISc Bangalore  \hspace{2.1em}
$^4$Fast Code AI
} 

\twocolumn[{ 
\renewcommand\twocolumn[1][]{#1} 
\maketitle 
\begin{center} 
    \vspace{-15pt} 
    \includegraphics[width=\textwidth ]{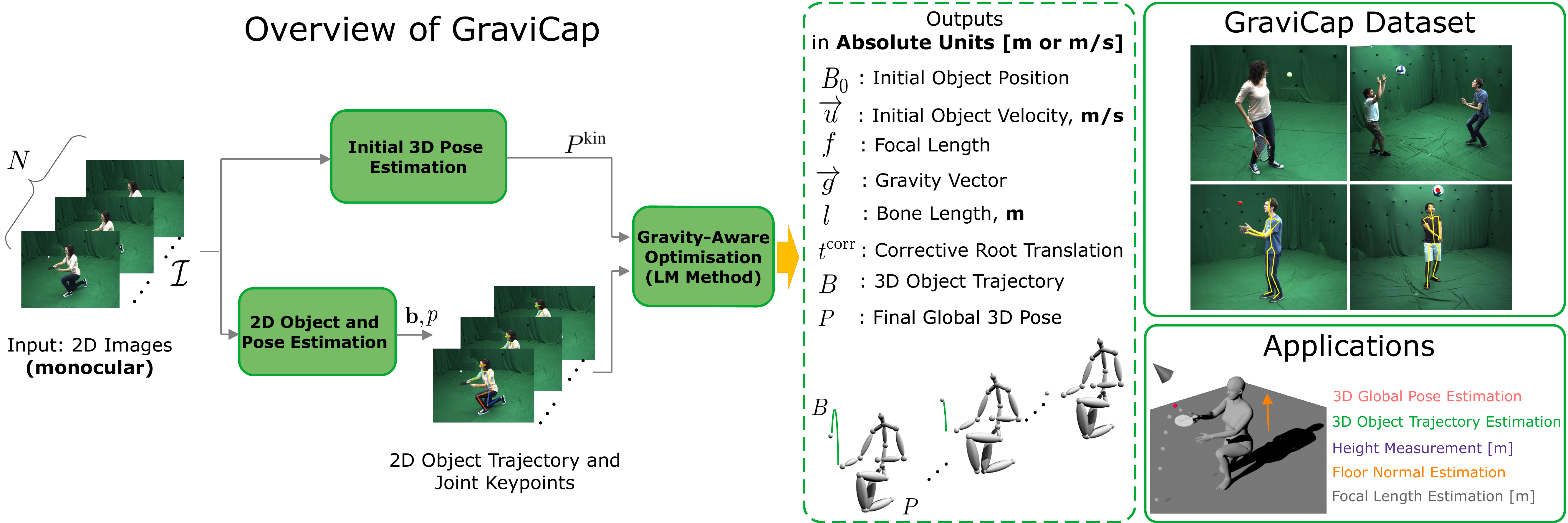} 
    \captionof{figure}{The proposed \GraviCap{} approach captures 3D human motions and 3D object trajectories from monocular RGB videos. 
    (Left:) Thanks to the physics-based constraints, we can disambiguate the scene's scale in the monocular setting and recover 3D human poses and 
    the trajectories in meters. 
    (Right:) We evaluate our method on a new real multi-view dataset with several subjects and activities. 
    } 
    \label{fig:teaser} 
\end{center} 
}] 

\begin{abstract} 
   This paper proposes \textnormal{GraviCap}, \textit{i.e.,} a new approach for joint markerless 3D human motion capture and object trajectory estimation from monocular RGB videos. 
   We focus on scenes with objects partially observed during a free flight. 
   In contrast to existing monocular methods, we can recover scale, object trajectories as well as human bone lengths in meters and the ground plane's orientation, thanks to the awareness of the gravity constraining 
   object motions. 
   Our objective function is parametrised by the object's initial velocity and position, 
   gravity direction and focal length, and jointly optimised for one or several free flight episodes. 
   The proposed human-object interaction constraints ensure geometric  consistency of the 3D reconstructions and improved physical plausibility of human poses compared to the unconstrained case. 
   We evaluate \textnormal{GraviCap} on a new dataset with ground-truth annotations for persons and different objects undergoing free flights. 
   In the experiments, our approach achieves state-of-the-art accuracy in 3D human motion capture on various metrics. 
   We urge the reader to watch our supplementary video. 
   Both the source code and the dataset are released; see 
   \url{http://4dqv.mpi-inf.mpg.de/GraviCap/}. 
\end{abstract}

\section{Introduction} 
    Markerless 3D human motion capture from a single monocular  RGB camera has many open challenges. 
    Although state-of-the-art methods have seen great progress  \cite{JMartinezICCV_2017, VNect2017, pavllo20193d, inthewild3d_2019, shimada2020physcap}, 
    they still hardly work for scenes showing non-trivial interactions of humans with the environment as most of them do not model environmental constraints or physical laws.  %
    Further, 3D reconstruction of humans interacting with objects from monocular imagery is scarcely explored, and only a few works were proposed to date \cite{li2019motionforcesfromvideo, zhang2020phosa}. 
    Most existing methods that consider interaction with the environment %
    impose geometric constraints to avoid incorrect interpenetrations \cite{Zanfir2018, Hassan2019, zhang2020phosa}. 
    They often exhibit strong jitter, implausible posture with unnatural body leaning and depth instabilities. 
    Recent physics-based methods for monocular 3D human pose estimation \cite{rempe2020contact,  shimada2020physcap} showed that explicit modelling of gravity and ground reaction forces (or friction) enables monocular reconstruction of humans of much higher biomechanical plausibility.  
    However, these methods do not model object interactions, and without a priori information about the human body, they cannot estimate posture and scene dimensions in absolute metric scale.

    In this paper, we make the following observation: Explicitly modelling physics 
    and actively encouraging a specific form of human-object interaction in the scene 
    \textit{enables improved 3D human and 3D object trajectory reconstruction in a metrically accurate way from a single monocular video}. 
    We consider scenarios when up to two persons are interacting with an  object and bringing it to a free flight (\textit{e.g.,} throwing or tossing). 
    Such scenarios are often observed in real everyday life  while practising sports or playing outdoor games. 
    We show that physics-based constraints allow us to obtain 3D estimates in the absolute units, which, otherwise, remains inaccessible for a  monocular setting when no strong prior assumptions about the scene can be made, such as known bone lengths. 

    Our core findings are that 1) Projectile motion constraints are sufficient to recover the 3D trajectory of an object undergoing free flight 
    from 2D object coordinates, assuming known camera frame rate and gravity vector; 
    2) Knowing the magnitude of the gravity and the focal length is sufficient to resolve the  scale of the observed scene in meters and orientation of the ground plane, assuming that the direction of the gravity vector %
    is opposite to the ground plane normal; 
    \hbox{3) Localising humans with} respect to the recovered 3D object  trajectory leads to improved 3D human motion capture. 
    See Fig.~\ref{fig:teaser} for an overview of our framework. 
    The inputs are 2D coordinates of the object's geometric centre 
    and 2D human joint locations, along with initial unconstrained kinematic 3D human poses. 
    After that, we then minimise the proposed  objective globally over multiple input frames and obtain 3D object trajectory over one or several free-flight episodes and improved 3D human poses. 
Summarised, the \textbf{contributions} of this work are as follows: 
\begin{itemize}[leftmargin=*]
\itemsep0em
    \item \GraviCap, \textit{i.e.,} the new approach for joint 3D capture of human motions and 
    trajectories of objects undergoing free flights   (Sec.~\ref{sec:approach}); 
    \item New types of human-object interaction constraints 
    improving the accuracy and physical plausibility of 3D human poses (Sec.~\ref{sssec:localisation_constraints}). 
    For the first time, these constraints allow 
    recovering camera-relative distances of moving and interacting  objects, including humans, in meters from a single monocular RGB camera;  
    \item A new dataset of human-object interactions for experimental evaluation, with ground-truth annotations of 3D human poses and object trajectories (Sec.~\ref{sec:dataset}). 
\end{itemize} 
We achieve state-of-the-art accuracy for global 3D human motion capture using different metrics in extensive experiments with the new dataset (Sec.~\ref{sec:experiments}). 
Our estimates look more physically plausible and temporally consistent  compared to results without human-trajectory localisation constraints. 
Moreover, the proposed constraints significantly improve absolute root translations. 
The source code of {\GraviCap} and the dataset 
are publicly available at  \url{http://4dqv.mpi-inf.mpg.de/GraviCap/}. 

\section{Related Work}\label{sec:related_work}

\noindent\textbf{Kinematic 3D Human Pose Estimation.} 
The accuracy of monocular 3D human pose estimation significantly progressed  during recent years. 
Most methods employ neural networks and can be classified into several categories. 
Some methods first estimate 2D poses %
in the input views and then lift them in the 3D space  \cite{Chen2017, JMartinezICCV_2017, Tome2017, Moreno-Noguer_2017, Fang2018, Dabral2019}, whereas several others estimate 3D joints directly from the images \cite{Tekin2016, mono-3dhp2017, Rhodin2018}. 
Several lifting algorithms build upon the principles of non-rigid structure from motion and rely on classical optimisation for the lifting step \cite{Zhou2015, Wandt2016, Kovalenko2019}. 
At the same time, weakly-supervised methods 
gain more and more attention, due to improved generalisability beyond the training datasets \cite{Dabral2018, WandtRosenhahn2019, Chen2019, Novotny2019}. 
Many other approaches combine regression of 2D joint locations or 3D joint depths \cite{Newell2016, VNect2017, Pavlakos2017, inthewild3d_2019}. 
Parametric body models provide strong priors on plausible shapes and poses, which can be leveraged for accurate human pose estimation \cite{Bogo2016, hmrKanazawa17, Pavlakos2018, Kocabas2020}. 
Even a stronger prior is a human mesh, and several recent methods show how to use it for tracking a single actor \cite{Habermann2019, Habermann2020, Xu2020}. 
In contrast to all approaches discussed so far, several other techniques %
generalise to scenarios with multiple subjects \cite{Dabral2019, Rogez2019, Moon_2019_ICCV_3DMPPE,  XNect_SIGGRAPH2020}. 
Several purely kinematic methods attempt to  estimate 3D human poses with absolute depths in the camera coordinate space  \cite{Moon_2019_ICCV_3DMPPE, shi2020motionet, XNect_SIGGRAPH2020}. 
All approaches reviewed so far consider geometric fidelity of the reconstructed motions and do not impose environmental constraints.
\noindent\textbf{3D Human Pose Estimation with Environmental Priors.} 
Hassan \textit{et al.}~\cite{Hassan2019} use 3D environmental scans to detect human-object collisions and improve kinematic 3D pose regression. 
Environmental constraints such as a common ground plane and volume occupancy exclusions %
are effectively applied in Zanfir  \textit{et al.}~\cite{Zanfir2018} for 3D human pose and shape estimation. 
Zhang \textit{et al.}~\cite{zhang2020phosa} jointly reconstruct humans and objects relying on geometric shape priors, 
both for humans and objects, as well as interactional vicinity. 
iMapper of Monszpart and colleagues \cite{Monszpart2019} jointly recovers  schematic 3D scene arrangements and human motions in a data-driven manner,  relying on a database of 3D human-object interactions for training. 
The authors show that motion patterns provide a strong cue about scene compositions, which, in turn, serve as priors for possible human motions. 
Similarly, we find in this paper that the physics-based cues associated with the object's motion caused by the  gravitation 
can better constrain human poses. 
Vondrak \textit{et al.}~\cite{vondrak2012video} capture 3D human motions by recovering 3D bipedal controllers that simulate motions observed in the videos. 
Li \textit{et al.}~\cite{li2019motionforcesfromvideo} 
simultaneously estimate 3D trajectories of human skeletal joints and an instrument (used by the person), as well as forces at contact positions (\textit{i.e.,} foot-floor and hand-object contacts). 
They observe that the instrument provides a reconstruction cue for hands in 3D (\textit{i.e.,} for their relative positioning in depth), and the hand positions provide a cue for the instrument's 3D position. 
In contrast, %
we focus on objects that can be released and move freely under 
gravity along a ballistic trajectory. 
Three recent methods constrain human motions with bio-physical plausibility constraints \cite{rempe2020contact, shimada2020physcap, PhysAwareTOG2021}. 
This allows significantly reducing unnatural body leaning, foot-floor penetration and jitter. 
We formulate physics-based constraints for objects and not directly humans. 
In contrast to all reviewed methods using environmental priors, we can disambiguate the scale of the scene and calculate the distances (\textit{e.g.,} bone lengths and 3D object trajectories) in meters. 
Moreover, 3D human poses estimated this way are more  physically-plausible compared to the initial kinematic estimates, thanks to our human-object localisation constraints. 
\noindent\textbf{Other Related Problems.} 
As a side effect, our \GraviCap{} can extract absolute bone lengths from  monocular videos. 
Several methods extract anthropometric measurements using 3D 
registration techniques and assume 3D human body scans as inputs \cite{Tsoli2014, Wasenmueller2015}, whereas we rely on 2D video inputs
only. 
Bieler \textit{et al.}~\cite{Bieler2019} use equations of classical mechanics for estimation of human height from videos of jumping people. 
We can estimate human height as a by-product of scale disambiguation and due to the reconstructed object motion.

Bhat \textit{et al.}~\cite{Bhat2002} show how to estimate rigid body's motion in free flight with a simulation that agrees with the image observations. 
Assuming a known shape of a small order of rotational symmetry allows estimating the initial position and velocity of the object, gravity direction and extrinsic parameters of the object relative to the camera. 
In contrast, we assume that 1) The target objects have infinite order of rotational symmetry (\textit{i.e.,} they are spherical) and 2) Their diameter is unknown. 
We show that these assumptions are sufficient to disambiguate the scene's scale. 

\section{Approach}\label{sec:approach} 
We now describe our \GraviCap{} approach for jointly recovering the object and human trajectories in the camera's frame of reference;  see Fig.~\ref{fig:teaser} for an overview. 

\subsection{Recovering the 3D Object   Trajectory}\label{ssec:objects_trajectory} 
First, we assume known camera focal length $f$ and a set of 2D observations of an object's ballistic trajectory $\bfb = \{b_1, b_2, \dots b_N\}$ extracted from images $\mathcal{I} = %
\{\mathcal{I}_1, \mathcal{I}_2, \dots, \mathcal{I}_N\}$, where $b_i = (x_i, y_i)$ is the object's position in image 
$i \in \{1, \hdots, N\}$. 
Our goal is to recover the object's 3D trajectory $B = \{B_1, B_2, \dots B_N\}$, where $B_i = (X_i, Y_i, Z_i)$ represents the object's position in the camera-relative 3D space. %
We call \textit{episode} one free flight event observed in a monocular video. 
We assume that once released, the only force influencing the motion of the object is gravity (there is no air resistance). 
This assumption allows us to parameterise %
$B$ using three parameters: The initial velocity  $\mathbf{\overrightarrow{u}} = (u_x, u_y, u_z)$, the object's initial  position $B_0$ and the gravity vector $\overrightarrow{g} = (g_x, g_y, g_z)$ as viewed in the camera's frame of reference. Given frame rate $r$, 
$B$ can be expressed using the equations of Newtonian dynamics as  
\begin{equation}
    B_i = B_0 + \overrightarrow{u}t + \frac{1}{2}\overrightarrow{g}t^2, 
\end{equation}
where $t = i / r$ is the time stamp in seconds corresponding to the frame $i$ from the beginning of the free flight. 
Next, assuming intrinsic camera parameters (focal length $f$ and principal point $c = (c_x, c_y)$) and gravity vector $\overrightarrow{g}$ are known, 
it is possible to reconstruct the 3D trajectory $B$ of the object from 2D observations $\bfb$. 
Under the pinhole camera model, the observed object's trajectory in the video can be explained as follows: 
\begin{equation}\label{eq:forward_model} 
\begin{aligned} 
    x_i = f \frac{X_i}{Z_i} + c_x, \; 
    y_i = f \frac{Y_i}{Z_i} + c_y, \; \forall i, \\ 
    \text{s.~t.}\;\;\sqrt{g_x^2 + g_y^2 + g_z^2} = 9.81\,m/s^2, 
\end{aligned} 
\end{equation} 
\begin{equation}\label{eq:forward_model_I} 
\begin{aligned} 
    \text{where}\; 
    \begin{cases} 
           X_i = X_0 + u_xt + \frac{1}{2}g_xt^2, \\ 
           Y_i = Y_0 + u_yt + \frac{1}{2}g_yt^2, \;\,\text{and} \\ 
           Z_i = Z_0 + u_zt + \frac{1}{2}g_zt^2. 
    \end{cases} 
    \end{aligned} 
\end{equation} 
The equation system \eqref{eq:forward_model} has $3N$ unknowns. 
Using the parametrisation with the ballistic trajectories  \eqref{eq:forward_model_I}, it reduces to six, \textit{i.e.,} three for the initial position $B_0 = (X_0, Y_0, Z_0)$ and three for 
$\mathbf{\overrightarrow{u}}$. 
Thus, \eqref{eq:forward_model} has a unique  solution when $N{>}2$ (for $N{=}3$, it has a  closed-form solution). 
We next consider two cases, \textit{i.e.,} when the direction of $\overrightarrow{g}$ is 1) known and when it is 2) unknown in \eqref{eq:forward_model}. 
In the first case, we assume that the direction of $\overrightarrow{g}$ is parallel to the $y$-axis and coincides with the flipped floor normal in the world coordinate system. 
This is highly relevant in practice, especially in artificial environments. 
In the second case, the orientation of the ground plane with respect to the camera remains unknown. 
Consequently, \eqref{eq:forward_model} contains three more unknowns and has a solution if $N{>}4$.  
At the same time, in both cases we assume that the magnitude of $\overrightarrow{g}$ is known and equals $9.81\,m/s^2$, which is a reasonable assumption. 
Even though $\overrightarrow{g}$ differs depending on the location on Earth, the  differences are insignificant and lie beyond the values which can improve the attainable precision in the 3D trajectory estimation from monocular images in our setting\footnote{ $\norm{\overrightarrow{g}} = 9.81\,m/s^2$ is close to the mean value of  $\norm{\overrightarrow{g}}$ on the surface of Earth, and $\norm{\overrightarrow{g}}$ differs not more than by ${\approx}0.7\%$ across locations.}. 
If both $f$ and $\overrightarrow{g}$ are unknown, \eqref{eq:forward_model} has ten unknowns which can be recovered with $N{>}5$, subject to proper initialisation (see comments on the `$f/Z$' ambiguity in Sec.~\ref{sec:discussion}). %
In practice, we use and recommend $N{>}10$ to obtain a better determined system (compared to 
$N{=}5$). 
A solution to such a system 
is less  sensitive to noise in the 2D measurements and quantisation effects. 
Table \ref{tab:algorithmic_modes} summarises the operational modes of \hbox{\GraviCap{}} and the corresponding sets of unknowns for the object's trajectory reconstruction. 

We solve \eqref{eq:forward_model} for $B_0$,  $\mathbf{\overrightarrow{u}}$ and, optionally, $f$ and  $\overrightarrow{g}$ by minimising the objective $E_b = E_b(B_0, \mathbf{\overrightarrow{u}}, f, \overrightarrow{g})$ in $\ell_2$-norm: 
\begin{align}\label{eq:main_trajectory_recovery} 
    \begin{split} 
    \arg \min_{B_0, \mathbf{\overrightarrow{u}}, f, \overrightarrow{g}} \sum_i \norm{ 
                        \begin{bmatrix} x_i \\ y_i \end{bmatrix} -          \begin{bmatrix}f\frac{X_i}{Z_i} + c_x \\                                  f\frac{Y_i}{Z_i} + c_y 
                        \end{bmatrix} 
                        }_2^2, 
    \end{split} 
\end{align} 
with $X_i$, $Y_i$ and $Z_i$ parameterised as in \eqref{eq:forward_model_I}. 
Note that the recovered 
$B_0$ and 
$\mathbf{\overrightarrow{u}}$ 
are in absolute units, \textit{i.e.,} $m$ and $m/s$, since  $\norm{\overrightarrow{g}}$ is expressed in $m/s^2$ and $f$ (if known) and  $t$ are expressed in meters and seconds, respectively. 

\noindent\textbf{Remark.} From \eqref{eq:forward_model}, we see that if there is no motion along $x$- (camera is observing a free fall and the free fall plane is parallel to the image plane) or $z$- (the free flight plane is parallel to the camera plane) axes, we can still recover the distances in absolute units, since gravity is affecting the $y$-component of the object's trajectory only. 
\vspace{5pt}

While the above formulation \eqref{eq:main_trajectory_recovery} is sufficient to recover the object's trajectory in ideal settings, the  accuracy of the estimated trajectory is sensitive to and is often  compromised by the observation noise. 
The sources of this noise can be multiple, including the missing and erroneous centre of gravity detections. 
We next show how the object's trajectory and human pose estimation can improve each other. 

\begin{table}[t] 
    \centering 
    \begin{tabular}{p{105pt}|p{42pt}|p{52pt}} 
        $\;$\textbf{mode / recovery of $\hdots$} 
        & $\;\;\,$\textbf{inputs} 
        & $\;$\textbf{unknowns}                       \\ \hline 
        3D object coordinates, scene scale (6 DoF)
        & $\bfb, \overrightarrow{g}$, $f$   
        & $\mathbf{\overrightarrow{u}}$, $B_0$    \\ 
        +gravity direction (9 DoF) 
        & $\bfb$, $\norm{\overrightarrow{g}}$, $f$
        & $\mathbf{\overrightarrow{u}}$, $B_0$, $\overrightarrow{g}$\\
        +focal length (10 DoF) 
        & $\bfb$, $\norm{\overrightarrow{g}}$ 
        & $\mathbf{\overrightarrow{u}}$, $B_0$, $f$, $\overrightarrow{g}$ \\ \hline 
        6 DoF + $f$ (7 DoF) 
        & $\bfb$, $\overrightarrow{g}$ & $\overrightarrow{u}$, $B_0$, $f$
         \\
        \hline 
    \end{tabular} 
    \caption{
    Different operational modes of \GraviCap{} for the 3D object trajectory recovery, 
    with the summary of inputs and unknowns. 
    Knowing $\mathbf{\protect\overrightarrow{u}}$, $B_0$, $f$ and  $\protect\overrightarrow{g}$ allows us to reconstruct $B_i$. 
    }\label{tab:algorithmic_modes} 
\end{table}

\subsection{Joint 3D Human-Object   Reconstruction}\label{ssec:joint_human_object} 
Associating the object's trajectory with the human's position provides additional %
constraints while also allowing us to estimate anthropometric information about one or multiple persons in the scene, thanks to the trajectory estimate in the absolute distance units. 
We first recover unconstrained kinematic estimates of 3D human skeleton $P^{\text{kin}} = \{P_1^{\text{kin}}, P_2^{\text{kin}}, \dots,  P_N^{\text{kin}}\}$, where $P_i^{\text{kin}} \in \mathbb{R}^{K\times3}$ and $K{=}16$ is the number of joints. 
We denote individual 3D $x$-, $y$- and $z$-components of each joint, indexed by $k \in \{1, \hdots, K\}$, using $P_{i, k, x}^{\text{kin}}$, $P_{i, k, y}^{\text{kin}}$ and $P_{i, k, z}^{\text{kin}}$, respectively. 
$P^{\text{kin}}$ can be either root-relative or also include an initial estimate of the root translation. 
In both cases, $P^{\text{kin}}$ is estimated separately from the object's 3D trajectory and, hence, is not provided in absolute coordinates and can be physically implausible. 
We use an off-the-shelf 2D pose estimator RMPE (\hbox{AlphaPose})~\cite{alphapose} to  extract 2D poses of the person $p = \{p_1, p_2, \dots p_N\}$, where $p_i \in \mathbb{R}^{K\times2}$, observed in the input images $\mathcal{I}$. 
The root-relative 3D poses $P^{\text{kin}}$ of the same person can then be retrieved using an off-the-shelf human pose estimation method like~\cite{VNect2017, Moon_2019_ICCV_3DMPPE, Dabral2018}. 
Those are either lifting methods 
(\textit{i.e.,} they operate on $p$) or direct regression approaches operating on $\mathcal{I}_i$. 
Since these techniques are monocular (RGB-based), 
they either predict 3D poses with a canonical skeleton or lack generalisability across different people (body variations). 

Our goal is thus to recover the bone lengths, $l = \{l_1, l_2, \dots, l_{K-1}\}$ of the subject, such that the corresponding root-relative 3D poses 
$\operatorname{s}(P^\text{kin}, l)$ are in the true metric space and 
agree with the anatomical lengths. The operator $\operatorname{s}(\cdot, \cdot)$ resolves the scale of $P^\text{kin}$ by  
rectifying  
the bone-lengths of $P^\text{kin}$ with the estimated bone-lengths $l$, such that the bone direction vectors are preserved. 
 Furthermore, we also estimate the corrective root translations $t^{\text{corr}} =  \{t_1^{\text{corr}}, t_2^{\text{corr}}, \dots, t_N^{\text{corr}}\}$ of the person from the  camera centre, where $t_i^{\text{corr}} = (t_{i, x}^{\text{corr}}, t_{i, y}^{\text{corr}}, t_{i,z}^{\text{corr}})$. 
Once the latter are available, the absolute (global)  camera-relative pose of the person can be recovered as $P = s(P^{\text{kin}}, l) + t^\text{corr}$. 

\begin{figure} 
    \centering 
    \includegraphics[width=\linewidth]{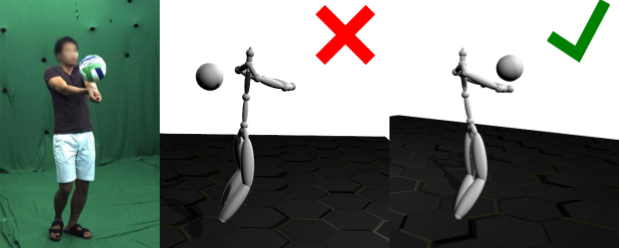} 
    \caption{Independently optimising human root translation and object's trajectory can result in incoherence between the two (middle). We rectify these artefacts by binding human and object positions with contact and mutual localisation constraints.
    } 
    \label{fig:contact_artifact} 
\end{figure}

We next assume that in the considered free flight episode, we know when the person is  holding the object and when the free flight  starts. 
This allows disambiguating the person's scale using the recovered trajectory $B$ in the absolute coordinates. 
Knowing that the object is in contact with the human is necessary. 
At the moment of contact, the human scale is the same as the trajectory scale (recall that under \textit{scale}, we mean the factor relating the relative and absolute distance units). 
If there is no contact, the usual ambiguity of the monocular setting along the depth axis still applies to the human and other parts of the scene. 
We recover the subject's poses 
with respect to the  camera by minimising $E_p =  E_p(l, t_i^\text{corr})$ in  $\ell_2$-norm: 
\begin{align} 
    \begin{split} 
    \arg \min_{l, t^\text{corr}} \sum_{i, k} %
    \norm{
                        \begin{bmatrix} p_{i,k}^x \\ p_{i,k}^y \end{bmatrix} -          \begin{bmatrix}f\frac{s(P_{i,k}^{\text{kin}}, l)[x] + t_{i, x}^{\text{corr}}}{s(P_{i,k}^{\text{kin}}, l)[z] + t_{i, z}^{\text{corr}}} + c_x \\                                   f\frac{s(P_{i,k}^{\text{kin}}, l)[y] + t_{i, y}^{\text{corr}}}{s(P_{i,k}^{\text{kin}}, l)[z] + t_{i, z}^{\text{corr}}} + c_y
                        \end{bmatrix} 
                        }_2^2, 
    \end{split}\label{eq:e_p_bl} 
\end{align} 
with operator $\cdot[\bullet]$ extracting $x$-, $y$- or $z$-component of a vector (alternative notation). 
In (\ref{eq:e_p_bl}), we have $2NK$ equations and $(3N{+}K)$ unknowns for root translations $P^r$ ($3N$ unknowns), bone lengths $l$ ($K{-}1$ unknowns) and the focal length $f$. 
However, the system of equations \eqref{eq:e_p_bl}, if considered independently from $E_b$ \eqref{eq:main_trajectory_recovery}, suffers from scale ambiguity because the bone lengths and root translations counteract each other. 
Furthermore, pose estimates, both 2D and 3D, are prone to errors due to model inaccuracies; especially, in the case of occlusions. 
Hence, the estimated object trajectory and the human pose are not guaranteed to be in agreement. 
It is thus natural to expect reconstructions where the object's release  position is far away from the point of contact 
like hands or feet, see Fig.~\ref{fig:contact_artifact}. 
We mitigate these issues in two ways. Firstly, we use a prior on the human bone lengths and impose a symmetry constraint. 
Secondly, we bind the person's absolute pose with the object's trajectory using two additional constraints discussed next. 

\subsubsection{Constraints on Human  Skeleton}\label{ssec:constraints_on_skeleton} 
The first constraint on the bone lengths $E_{bl}(l)$ ensures that the estimated bone lengths are close the average human bone lengths $\bar{l} = \{\bar{l}_k\}$. 
For this, we use the average bone lengths collected from the 
MPI-INF-3DHP~\cite{mono-3dhp2017} dataset: 
\begin{equation}\label{eq:bl_prior} 
    \arg \min_{l}\,E_{bl}(l) = \arg \min_{l}\sum_{k}^{K-1} \norm{l_k - \bar{l}_k }_2^2. 
\end{equation} 
Additionally, we ensure that the recovered $l$ 
are left-right symmetric using the 
symmetry constraint $E_s(l)$: 
\begin{equation}\label{eq:Es} 
    \arg \min_{l}\,E_s(l) = \arg \min_{l}\sum_{i,j \in \mathcal{S}} \norm{ l_i - l_j }_2^2,  
\end{equation} 
where $\mathcal{S}$ is the set of indices of symmetric bones; 
$E_s$ corrects asymmetries %
observed in the initialisations $P_i^{\text{kin}}$, thereby improving the plausibility and accuracy of 
$P_i^{\text{kin}}$.

\subsubsection{Human Contact and Localisation Constraints}\label{sssec:localisation_constraints} 

The contact term $E_c(P)$ expresses the prior assumption that the object is thrown or caught by the person, \textit{i.e.,} it ensures  that
the 3D positions of the object and the corresponding body joints at the moment of contact are close to each other: 
\begin{equation}\label{eq:contact_term} 
\small
    \arg \min_{\overrightarrow{u}, \overrightarrow{g}, B_0} E_c(P) = \arg \min_{\overrightarrow{u}, \overrightarrow{g}, B_0} \sum_{(c, t) \in \mathcal{C}} \norm{P^c_t -  B_t }_2^2, 
\end{equation} 
where $\mathcal{C}$ denotes the set of joints in contact with the object at time $t$, and $P^c_t$ are the 3D coordinates of these joints. 

Although \eqref{eq:contact_term} binds the object's trajectory with the human's absolute position at the points of contact, it does not explicitly associate the two for the rest of the frames. 
Also, it does not generalise well to settings where the  object is not close to the body at the contact points (\textit{e.g.,} when hitting a ball with a tennis racket). 

\begin{figure}[t!]
    \centering 
    \includegraphics[width=0.9\linewidth]{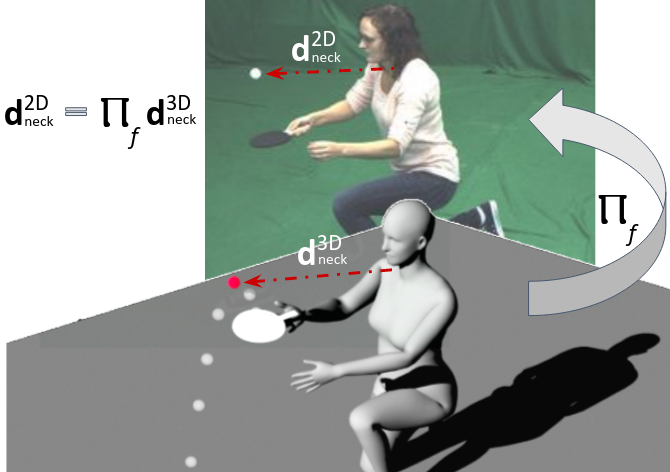} 
    \caption{Illustration of the principle behind the human-object localisation term $E_m$ for the neck joint and the second frame of the free flight episode. 
    $E_m$ ensures that the points along the vector connecting the object and the neck joint in 3D ($\mathbf{d}^{3D}_{neck}$) must project to the corresponding points in the 2D projection of the line ($\mathbf{d}^{2D}_{neck}$) under perspective projection. In addition to the neck joint, we also use hips, spine and shoulders to compute $E_m$.} 
    \label{fig:localisation_term} 
\end{figure}

To address this, we add a \textit{mutual} human-object localisation term $E_m$ to the objective, which ensures that the 3D vectors between the object's position at frame $i$ and the person's torso joints---when projected to the image---produce the corresponding observed vectors in the image plane between the object and human joints. 
This is possible and results in improved 3D human motion capture because  the 3D object's trajectory is smooth and can be estimated highly accurately; we can reliably localise the human with respect to it. 
For $E_m = E_m(B_0, \mathbf{\overrightarrow{u}}, f, \overrightarrow{g}, l, t^\text{corr})$, we choose the torso joints (pelvis, spine, neck, shoulders), because of their stable nature with respect to the camera-relative translation: 
\begin{equation}\label{eq:e_m} 
\small 
\begin{aligned}
    \arg \min_{B_0, \mathbf{\overrightarrow{u}}, f, \overrightarrow{g}, l, t^\text{corr}} \sum_i\sum_{j\in \mathcal{T}}\sum_m^M \norm{\mathbf{d}^{2D}_{i,j,m} - \Pi_f(\mathbf{d}^{3D}_{i,j,m})}_2^2, \\ 
\end{aligned}
\end{equation} 
\begin{equation}\label{eq:e_m_} 
\begin{aligned}
    \text{where}\quad& d^{2D}_{i,j,m} = p_{i,j} + \frac{m(b_{i,j} - p_{i,j})}{M}, \; \text{and} \\ 
    & d^{3D}_{i,j,m} = P_{i,j} + \frac{m(B_{i,j} - P_{i,j})}{M}. 
\end{aligned}
\end{equation} 
where $\mathcal{T}$ is the set of torso joints and $\mathbf{d}^{2D}_{i,j,m}$ and $\mathbf{d}^{3D}_{i,j}$ are the vectors between object and torso joint $j$ at frame $i$ in 2D and 3D, respectively.
Note that we optimise for 3D vectors guided by their reprojections in the image plane. 
Such optimisation also influences (in most cases, improves the accuracy of) the corresponding joints $j$. 
In practice, we uniformly sample $M$ points in $\mathbf{d}^{2D}_{i,j}$ and $\mathbf{d}^{3D}_{i,j}$, and penalise the projection error.
See Fig.~\ref{fig:localisation_term} for  illustration of this principle. 

\subsubsection{Multi-Episodes and Multi-Person Settings} 

So far, we have been focusing on a single person and a single episode, \textit{i.e.,} the case of a single ballistic trajectory. 
We also propose a variant of our method that can handle multi-episodes with consecutive ballistic trajectories and the same person. 
We assume that $f$, $\overrightarrow{g}$ and the human bone lengths $l$ are  constant over the multi-episodes, whereas $B_0$ and  $\mathbf{\overrightarrow{u}}$ are individual for every episode. 
To stitch the multiple trajectories coherently, we propose an additional constraint $E_{co}$, that penalises the differences between the initial object's position in the current episode and its last position in the previous episode using $\ell_2$-norm. 
Furthermore, our method allows for two-person (multi-) episodes (\textit{e.g.,} two persons throwing an object at each other) with minimal changes. 
To that end, we modify $E_p$ \eqref{eq:e_p_bl} and $E_m$ \eqref{eq:e_m} to account for the projection losses with both persons in the scene. 
Likewise, the contact loss $E_c$ \eqref{eq:contact_term} is altered to account for contacts with different persons at different time instants; see the supplementary material for details. 

\subsection{Joint Energy Optimisation}\label{ssec:joint_optimisation} 

The joint objective term of \GraviCap{} can now be expressed as a sum of seven energy terms: 
\begin{equation}\label{eq:joint_opt}
\small 
     E =  E_p + \lambda_bE_b + \lambda_cE_c + \lambda_mE_m + \lambda_sE_s + \lambda_{co}E_{co} + \lambda_{bl}E_{bl}, 
\end{equation}
with weights $\lambda_b, \lambda_c, \lambda_m$,  $\lambda_s$, $\lambda_{bl}$ and $\lambda_{co}$ balancing the individual terms. 
As mentioned in Sec.~\ref{ssec:objects_trajectory}, \GraviCap{} can work in four modes with 6 DoF ($\mathbf{\overrightarrow{u}}$, $B_0$), 7 DoF ($\mathbf{\overrightarrow{u}}$, $B_0$, $f$) 9 DoF ($\mathbf{\overrightarrow{u}}$, $B_0$, $\overrightarrow{g}$) and 10 DoF ($\mathbf{\overrightarrow{u}}$, $B_0$, $\overrightarrow{g}$, $f$) for the object's trajectory. 
The total number of unknowns in \eqref{eq:joint_opt} is $3N{+}K{+}9$ for one person and one episode. 
At the end of this optimisation, our method can reconstruct the object's absolute 3D trajectory along with the human's absolute root positions for all the time steps.
We solve for the unknowns in \eqref{eq:joint_opt} using  Levenberg-Marquardt~\cite{levenberg-marquardt} with $\lambda_b{=}1.0, \lambda_p{=}1.0,  \lambda_c{=}0.1, \lambda_m{=}0.5, \lambda_{co}{=}0.1, \lambda_{bl}{=}0.1$ and $\lambda_s{=}0.01$. 
The values of balancing terms are chosen empirically using a hyper-parameter sweep.
Note that optimising first  \eqref{eq:main_trajectory_recovery} disjointly from  \eqref{eq:joint_opt} produced worse results. 

\begin{table} 
 \footnotesize
 \scalebox{0.95}{
 \begin{tabular}{ c|c|c|c|c|c|c|c|c|c|c}\hline
            & S1 & S2 & S3 & S4 & S5 & S6 & S7 & S8 & S9$^*$ & $\sum$  \\ \hline \hline 
      Ball  &  R  & PP   & PP   &  V  &  V  &  T  & V   &  T  &  V  & -         \\
      $L$, sec.  &  $8$  & $19$  & $26$ &  $14$  &  $21$  & $31$ & $21$ & $28$ & $21$  & $193$       \\
      $\sum$ ME   & $9$   &  $3$  & $6$ & $9$   & $9$   & $12$  & $6$   & $6$   &  $3$  & $63$       \\ 
      $\sum$ E    &  $12$  &  $21$   &  $45$   &  $18$   &  $54$   &   $30$   &  $30$   & $24$   & $18$   &  $252$   \\ \hline 
\end{tabular}
} \caption{Our dataset contains eight sequences with single subjects (S1--S8) and one sequence with two subjects (S9$^*$). 
Contents rowwise: ($1$) Object type, either rubber (R), ping pong (PP), volley- (V) or tennis (T) ball; ($2$) duration $L$ in seconds; ($3$) number of multi-episodes (ME); ($4$) number of episodes (E). 
}
\label{tab:recordings}
\end{table}

\subsection{Implementation Details}\label{ssec:implementation_details} 
For every input video, we follow a three-step approach:  Retrieving the kinematic human poses $P^{\text{kin}}$,  tracking the 2D object's trajectory $\bfb$, and finally,  minimising \eqref{eq:joint_opt}. 

\noindent \textbf{Estimating Human Poses.} 
We estimate the initial 3D positions of human skeleton joints using the real-time VNect method~\cite{VNect2017} for the single-person case and XNect~\cite{XNect_SIGGRAPH2020} for the multi-person setting. 
These methods provide absolute 3D positions 
in camera coordinates which serve as a reasonable initialisation for \GraviCap{}. 
For 2D joint positions, we use AlphaPose~\cite{alphapose}. 
To accommodate a differing number of joints across methods, we  consider the skeleton structure of MPII 2D pose dataset \cite{Andriluka2014} with $K{=}16$ joints. 

\noindent\textbf{Object Tracking Method.} 
The object tracklets are retrieved using OpenCV's off-the-shelf CSR Tracker~\cite{Lukezic_CVPR_2017}. 
For initialising the tracker, we localise the object in the first frame by performing template matching with a reference image. If an object detector exists for the target object (\textit{e.g.,} basketball), we recommend using it for the localisation. 
For the multi-episode setting, we detect the switch of  episodes based on the sudden change of the object's 2D  velocity direction. 
Since this change also happens during the movement along the ballistic trajectory (though much slower), we use a threshold on the velocity direction to distinguish the episodes  (\textit{cf.}~our supplement). 

\begin{table*}[!t]
\small
\centering
\begin{tabular}{l l|c c c c c c c c| c}
\hline
            &  & S1 & S2 & S3 & S4 & S5 & S6 & S7 & S8 & Avg\\ 
\hline
\multirow{3}{*}{\shortstack{global root positions,\\ MPE [mm] $\downarrow$}} 
& PhysCap~\cite{shimada2020physcap} & 431.1 & 225.6 & \textbf{232.3} & 239.2 & 223.1 & 446.5 & 358.8 & 420.28  & 309.7 \\
& VNect~\cite{VNect2017}  &  413.3 & 136.6 & 239.8 & 175.0 & \textbf{153.9} & 354.4 & 324.0 & 438.0 & 262.0\\
& Ours (10 DoF) & \textbf{219.7} & \textbf{111.3} & 234.8 & \textbf{\textit{166.9}} & \textbf{\textit{157.4}} & \textbf{\textit{332.1}}  & \textbf{\textit{187.7}} & \textbf{\textit{377.4}}  & \textit{\textbf{224.5}}\\
& Ours (9 DoF) & \textbf{\textit{411.9}} & \textbf{\textit{132.3}} & \textbf{\textit{232.7}} & \textbf{118.9} & 157.6 & \textbf{149.9}  & \textbf{135.7} & \textbf{352.4} & \textbf{191.3}\\
\hline
\multirow{3}{*}{\shortstack{$\;\,$root-relative poses,\\$\;\,$ MPJPE [mm] $\downarrow$}} 
& PhysCap~\cite{shimada2020physcap} & 126.3 & 128.6 & \textbf{94.8} & \textbf{\textit{129.1}} & 116.9 & 138.4 & 148.8 & 131.9  & 125.2\\
& VIBE~\cite{Kocabas2020} & 126.4 & \textbf{114} & 104.81 & \textbf{99.3} & \textbf{105.3} & 126.9 & \textbf{\textit{132.05}} & \textbf{105.3}  &  \textbf{113.2}\\ 
& VNect~\cite{VNect2017}  &  127.7 & 131.0 & 125.33 & 150.2 & 128.6 & 134.4 & 143.8 & 140.8 & 134.0 \\
& Ours (10 DoF) & \textbf{\textit{120.7}} & \textbf{\textit{119.6}} & 109.9 & 130.8 & 120.03 & \textbf{\textit{126.3}} & 140.4 & 134.6 & 124.4\\
& Ours (9 DoF) & \textbf{119.2} & 127.6 & \textbf{\textit{108.9 }}& 144.7 & \textbf{\textit{113.1}} & \textbf{125.8} & \textbf{\textit{130.7}} & \textbf{\textit{131.2}}  & \textit{\textbf{122.8}}\\
\hline
\multirow{2}{*}{\shortstack{$\;\;\;\,$bone lengths ($l$),\\$\;\;\;\,$MAE [mm] $\downarrow$}} & 
VNect\cite{VNect2017}  & 71.4 & 63 & 83.7 & 79.0 & 79.9 & 80.3 & \textbf{\textit{76.3}} & 83.8 & 78.4 \\
& Ours (10 DoF) & \textbf{64.0} & \textbf{\textit{52.6}} & \textbf{62.2} & \textbf{60.7} & \textbf{\textit{61.8}} & \textbf{58.5} & 84.4 & \textbf{64.6}  & \textit{\textbf{63.3}}\\
& Ours (9 DoF) & \textbf{\textit{64.3}} & \textbf{48.3} & \textbf{\textit{65.1 }}& \textbf{\textit{64.8}} & \textbf{59.5} & \textbf{\textit{65.8 }}& \textbf{60.2} & \textbf{\textit{65.1}} & \textbf{61.1}\\
\hline
\multirow{2}{*}{\shortstack{$\;$3D object positions \\$\;\,$($B_i$), MPE [mm] $\downarrow$}} & 
10 DoF & 451.0 & 309.0 & 760.3 & 304.0 & 476.8 & 372.3 & 280.6 & 470.7 & 445.5\\
& 9 DoF & 396.4 & 509.0 & 749.3 & 310.9 & 509.3 & 463.7 & 229.1 & 493.6 & 482.9\\
\hline
\multirow{2}{*}{\shortstack{gravity direction  ($\overrightarrow{g}$),\\cosine similarity $\uparrow$} } & 10 DoF & 0.927 & 0.977 & 0.954 & 0.949 & 0.99 & 0.99 & 0.951 & 0.975 &  0.972\\
& 9 DoF & 0.923 & 0.959 & 0.956 & 0.972 & 0.984 & 0.985 & 0.977 & 0.970 & 0.972\\
\hline
\end{tabular}
\caption{
Comparisons of various 3D errors on the new dataset (Sec.~\ref{sec:dataset}) with human-object interactions. 
Note that VIBE \cite{Kocabas2020} outputs root-relative poses only and, hence, cannot compete in global estimations. 
The \textbf{bold}/\textit{\textbf{italicised bold}} font denotes 
the best/second-best number. 
The last column provides the frame-weighted 
averages per sequence. 
`$\downarrow$'(`$\uparrow$') stands for `the lower (the higher) the better'. 
}\label{tab:pose}
\end{table*}

\section{Dataset for 3D Human-Object  Recovery}\label{sec:dataset} 
To evaluate the performance of our approach and establish an evaluation benchmark for future works, we record a new dataset with four subjects performing a variety of activities with four ball types. 
The dataset includes eight sequences with a single person and one additional sequence with two persons. 
For each sequence, we provide three synchronised videos of the scene, ground-truth intrinsic and extrinsic camera parameters, ground-truth  human poses (both 2D, for each view, and 3D) and ground-truth object  trajectories (both 2D, for each view, and 3D). 
For tracking 3D human joints, we use multi-view a markerless motion capture system \cite{captury} with $101$ camera views. Ground-truth object trajectories are recovered using triangulation. 
Each sequence contains several multi-episodes, \textit{i.e.,} consecutive sets of observed free flights. 
See Table \ref{tab:recordings} for the summary. 

\section{Experiments}\label{sec:experiments} 
To demonstrate the quality of the estimated 3D motions, we compare our results with existing state-of-the-art methods that estimate human poses in camera-relative space: 
VNect \cite{VNect2017}, MotioNet \cite{shi2020motionet} and  PhysCap \cite{shimada2020physcap}. 
Additionally, we include a recent method VIBE \cite{Kocabas2020} %
for root-relative pose estimation, which currently achieves state-of-the-art accuracy in this category. 
Note that VIBE is not a competing method since it cannot estimate global root translations. 
The pre-trained models that the authors provide are used for the comparisons. We report the root-relative Mean Per Joint Position Error (MPJPE), the Mean Position Error (MPE) of the camera-relative root translation, as well as smoothness error of the estimated 3D poses  \cite{shimada2020physcap}. 
We also evaluate the accuracy of the ground plane orientation estimation and report the cosine similarity between the ground-truth and estimated gravity direction vectors. 
Since \GraviCap{} also estimates bone lengths, we report Mean Absolute Error (MAE), \textit{i.e.,} mean of the per-bone absolute differences between the ground-truth and predicted lengths. 
We also report the estimated heights of subjects from the new dataset. 
Finally, we evaluate the accuracy of the object's trajectory using MPE. 

All experiments are performed on a system with 32GB RAM and AMD Ryzen ThreadRipper CPU, under operating system Ubuntu $18.04$. 
The method is implemented in SciPy \cite{2020SciPy-NMeth}. 
The runtime of the optimisation depends on the number of episodes in the sequence and ranges from three seconds for a single episode to roughly five minutes for a multi-episode with twelve trajectories. 

\subsection{Quantitative Results} 
\begin{figure*}[th!] 
    \centering 
    \includegraphics[width=\textwidth]{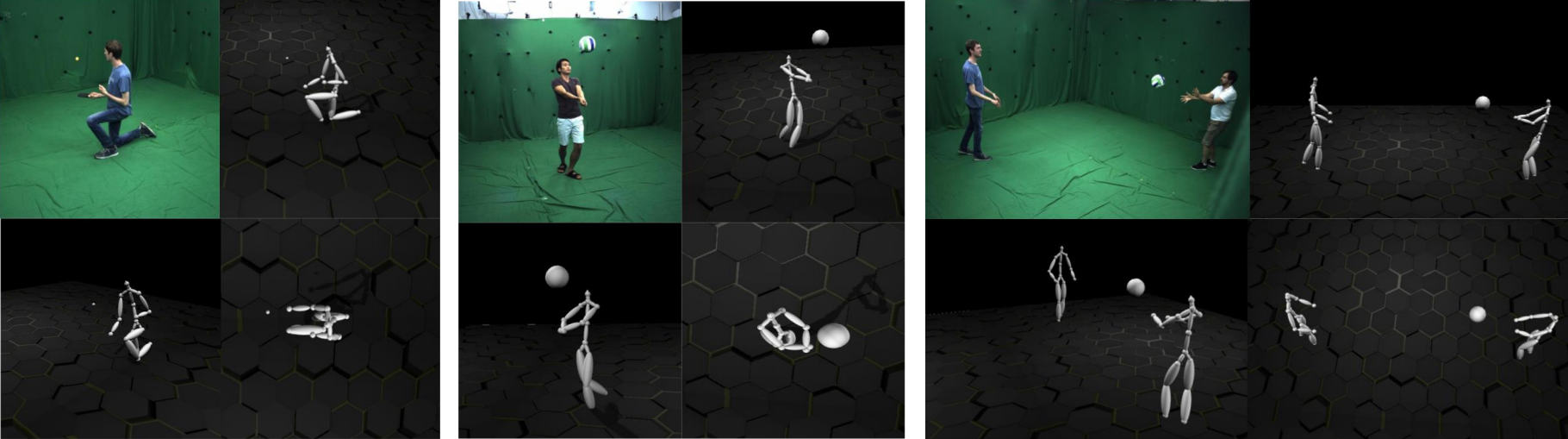} 
    \caption{3D human-object reconstructions of \GraviCap{} (from left to right: S2, S4 and S9). 
    In each image set out of three, we show the input view (top-left) and our reconstructions from three arbitrary views in 3D.
    See the supplementary video for the dynamic visualisations. 
    } 
    \label{fig:results} 
\end{figure*}

Table~\ref{tab:pose} summarises the comparisons of 3D human motion capture. 
9 DoF refers to the setting in which $B_0, \mathbf{\overrightarrow{u}}$ and $\overrightarrow{g}$ are unknown. 
In the experiments with 10 DoF, focal length $f$ is also unknown (however, VNect uses the ground-truth $f$), \textit{cf.}~Table \ref{tab:algorithmic_modes} with the summary of operational modes. 
We observe significant improvements in global root  translation estimation over VNect \cite{VNect2017} and PhysCap \cite{shimada2020physcap}, \textit{i.e.,} we outperform both methods in seven cases out of  eight. 
Note that MotioNet \cite{shi2020motionet} is not able to produce reasonable estimates, and we do not include it in Table~\ref{tab:pose}. 
All tested algorithms show MPJPE for root-relative 3D joint positions in the comparable ranges. 
While VIBE \cite{Kocabas2020} shows the lowest error for the root-relative poses overall, it cannot estimate global root translations. 
Note that only our method additionally estimates the floor normal, whereas other methods either require it as input or are agnostic to it. 
Fig.~\ref{fig:results} provides a few visualisations of the results obtained by \GraviCap{}. 
In bone length estimation, \GraviCap{} outperforms VNect in all cases.
Note that VNect and PhysCap implicitly assume a known average human height of a pre-defined skeleton. 
The second-lowest row block of Table~\ref{tab:pose} summarises the accuracy of the 3D object's  trajectories estimation (only our method can estimate those; the numbers are provided for future reference). 
\GraviCap{} can also estimate 
gravity directions. 
This is advantageous since we can obtain the floor normal,  
assuming 
that the gravity vector is perpendicular  
to the ground plane. 
We report cosine similarity between the ground-truth gravity direction and the estimated one in Table~\ref{tab:pose} (the lowest row block). 
As can be seen, our algorithm estimates highly accurate  gravity direction given only a monocular video. 

3D MPJPE considered in isolation can hide artefacts in the reconstructed motions such as jitters, which has been recently demonstrated \cite{shimada2020physcap}. 
Smoothness in motions is thus an important  criterion of physical plausibility. 
Therefore, we report smoothness error $e_{smooth}$ proposed in \cite{shimada2020physcap} in Table \ref{tab:smooth}. 
Our algorithm outperforms VNect and VIBE on this metric, thanks to mutual localisation term \eqref{eq:e_m} and stable bone length estimation. %
PhysCap performs best because of the explicit physics model and the assumption of known body mass distribution and subject's height, %
unlike our method. 

\vspace{10pt}

\noindent\textbf{Human Height Estimation.} We recover the subjects' heights from the estimated scale and root translations. 
We first calculate the head-toe distance $h_{px}$ in the image with the subject in the upright pose using AlphaPose \cite{alphapose, Bieler2019}. 
Since AlphaPose estimates only the head centre and ankles instead of head-top and feet/heels, we scale up the estimated height with a factor of $1.17$, as suggested in Bieler \textit{et al.}~\cite{Bieler2019}. 
Finally, we compute the actual height of the person as $h_{3d} = \frac{t^{corr}_z}{f} h_{px}$. 
We observe an average height error of $6.75\,cm$  across all four subjects of the new dataset. 

\begin{table}[h!]
\centering
\scalebox{0.9}{ 
\begin{tabular}{l|c|c|c|c}
    \hline
     & Ours & VNect~\cite{VNect2017} & VIBE~\cite{Kocabas2020} & PhysCap$^*$~\cite{shimada2020physcap}  \\
     \hline
$e_{smooth}$ & \textbf{10.74} & 11.35 & 11.32 & \textit{7.72}\\
\hline
\end{tabular}
} 
\caption{\label{tab:smooth}Comparisons using $e_{smooth}$. 
We reduce the jitter compared to VNect 
owing to the mutual-direction constraint and more accurate bone-length estimation. 
`$^*$' indicates the assumption of known body mass distribution and ground plane orientation. 
}
\end{table}

\subsection{In-the-Wild Experiments} 
In addition to the quantitative analysis on the new dataset, we test our method on in-the-wild settings (\textit{e.g.,} practising basketball and shot put); see our supplement. 

\section{Discussion and Limitations}\label{sec:discussion} 
Joint 3D human motion capture and 3D trajectory reconstruction of objects in free  flights is a new problem in computer vision. 
Our core method addresses it without relying on 3D training data; a learning-based approach is used only 
to initialise human poses. 
Yet, \hbox{\GraviCap{}} performs close to the state-of-the-art learning-based methods for root-relative human pose estimation \cite{Kocabas2020}, when compared in the accuracy of root-relative poses. 
Regarding absolute poses, we significantly outperform 3D motion capture methods estimating  camera-relative root translations \cite{VNect2017, shimada2020physcap, Moon_2019_ICCV_3DMPPE}. 
At the same time, our estimates of absolute bone lengths are the most accurate among all methods (if we assume average human height to convert normalised outputs to metric values for the competing techniques). 

Note that we use VNect for the initialisation, which is not state-of-the-art. 
This shows that the final superior accuracy is reached thanks to the proposed energy terms and awareness of the law of gravitation. 
In the absence of camera intrinsics, we either need well-initialised absolute root translations, or 
we can only estimate the scene's scale in most cases, due to the `$f/Z$' ambiguity in  \eqref{eq:forward_model}. 
Although the system is largely automatic, our framework strongly relies on the object's bounding box in the first frame and the hand-object contact  joint detections, which can be difficult to obtain without specialised methods (see Sec.~\ref{ssec:implementation_details}). 

To be applied on the Moon or Mars, \GraviCap{}  would require the corresponding local $\norm{\overrightarrow{g}}$. 
Our method assumes, per default, that humans hold objects in their hands. 
If this assumption is not fulfilled, it requires a prior on the type of the used instrument which propagates forces exerted by the human. 
Furthermore, since \GraviCap{} is a lifting approach, its accuracy depends on 2D human poses and 2D object detections, similar to several other monocular 3D human motion capture methods \cite{VNect2017, shimada2020physcap}. 

\section{Conclusion} 
We introduced \GraviCap{} and showed experimentally that given only a monocular video, in which humans interact with objects and bring them to free flights, it is possible to recover distances in meters (bone lengths and camera's focal length), the orientation of the ground plane relative to the camera, as well as significantly improve the initial kinematic human pose estimates reaching state-of-the-art accuracy. 
This changes the way how we think about the young  subfield of joint 3D human-object reconstruction 
and opens up many avenues for future research. 

\noindent\textbf{Acknowledgements.} 
This work was supported by the ERC consolidator grant 4DReply (770784). 

\small{
\bibliographystyle{ieee_fullname}
\bibliography{egbib}
}

\newpage 
\begin{center}
\textbf{{\Large Supplementary Material}} 
\end{center}
\appendix 

This supplementary material contains further details on \hbox{\GraviCap{}}. 
We also provide a supplementary video with further analysis of our method, in-the-wild 3D reconstructions and qualitative comparisons with other methods. 
\section{In-the-Wild Results}
To verify the accuracy of visual metrology achievable by our method, we test it on several real-world scenarios where the distance references are available. 
\begin{figure}[h]
    \centering
\includegraphics[width=\linewidth]{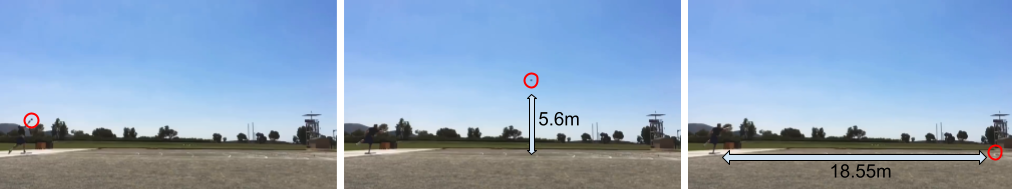}
    \caption{Shot put sequence}
    \label{fig:shotput}
\end{figure}

\noindent \textbf{Professional Shotput Throw:} A professional shot put thrower can throw in the range of $20$ meters. The image in Fig.~\ref{fig:shotput} is a reference to a throw by the world record holder Ryan Crouser\footnote{\url{https://www.youtube.com/watch?v=TZANFlvsXv4}}. 
Since the person is extremely blurred in the clip, we test our method with only object-related constraint  $E_b$, while ensuring that the magnitude of gravity vector is $9.81\,m/s^2$. 
Our method estimates the throw to be $18.557$ $m$ long. Further, we estimate the maximum point of the object's trajectory to be $5.46$ $m$.
Finally, the estimated gravity direction indicates an upward tilt of $13^\circ$ of the camera. 

\begin{figure}[t!]
    \centering
\includegraphics[width=\linewidth]{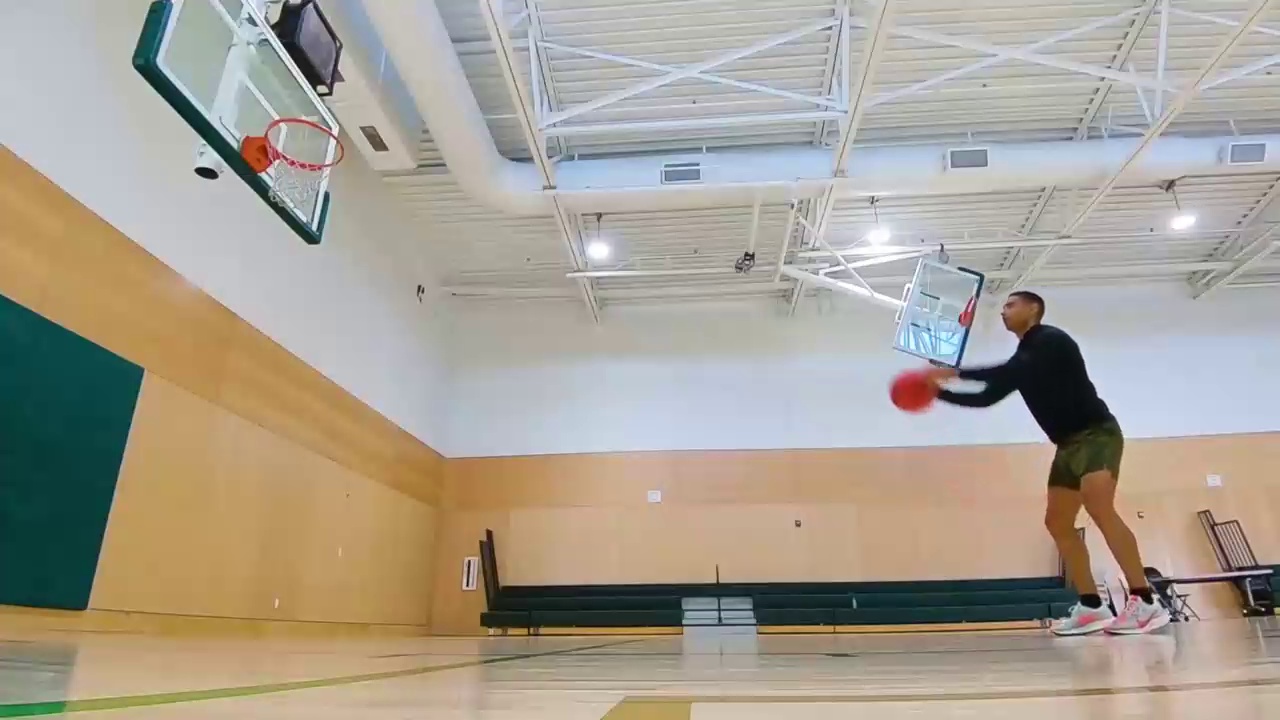}
    \caption{Basketball throw sequence from \url{https://www.youtube.com/watch?v=BKIOqbx3sbU}.}
    \label{fig:basketball}
\end{figure}

\noindent \textbf{Basketball Throw:} We measure the final position of the ball in the trajectory of the throw depicted in Fig.~\ref{fig:basketball}. We note the absolute $y$-position of the ball when it touches the hoop and compare it with the absolute $y$-position of the feet (as estimated by VNect). The difference between the two gives us an estimate of the height of the hoop. The trajectory estimates show a height of $3.03$ $m$, which is close to the actual height of the hoop ($3.05$ $m$). 

\begin{table}[h] 
    \vspace{-5pt} 
    \footnotesize 
    \centering 
    \begin{tabular}{l l l l l l} 
        & GT & $\sigma=10$ & $\sigma=30$ & $\sigma=50$ & $\sigma=100$ \\ \hline
        Pose, 6 DoF & 11.7 & 23.2 & 39.9 & 88.3 & 227.2 \\
        Pose, 7 DoF & 8.9 & 13.4 & 31.1 & 69.8 & 224 \\
        Pose, 10 DoF & 26.3 & 60.87 & 126.5 & 150.4 & 225.0 \\
        \hline
        Object & 12.4 & 76.3 & 134.4 & 155.8 & $>$400
    \end{tabular} 
    \caption{Comparing the effect of adding Gaussian  noise to the ground-truth 3D poses and 2D object trajectories on root translation predictions. 
    The unit of $\sigma$ is $mm$ for poses and pixels for 2D object trajectories. %
    }  
    \vspace{-15pt} 
    \label{tab:dof_exp} 
\end{table} 

\section{Noise Sensitivity Analysis} 
To test the sensitivity of \GraviCap{} to noise, we perform a sensitivity analysis and summarise the results in Table ~\ref{tab:dof_exp}. 
As expected, the performance is affected by strong 2D object trajectory perturbations, as high $\sigma$, \textit{i.e.,} the standard deviation of Gaussian noise, disrupts its parabolic nature. 

\section{Detection of Trajectory Breaks}
To detect an episode switch in multi-episode sequences, we traverse the 2D trajectory with a sliding window of five frames. 
During each slide, we measure the position difference between the positions of the objects  in adjacent frames. 
For a switch to have occurred, the direction of position differences of the first half must be opposite to that of the second half. 
This, however, is not sufficient for detecting a switch, as the same happens when the object reaches its peak. 
To confirm a switch, we measure the change in magnitude of velocity during the inversion. 
We set a threshold of $10$ pixels per frame to confirm the episode switch.

\section{Handling Multi-Episodes and Two Persons} 
\noindent \textbf{Multi-Episodes:} Whenever we have a multi-episode sequence, we perform joint optimisation on all the episodes. 
This leads to coherent reconstruction in the sense that the trajectory is continuous and jitter-free.
For this, we impose the continuity constraint, $E_{co}$, that ensures that the last position of the previous episode is the same as the first position of the current episode. Specifically, if $i=2,3,\dots$ refers to an episode in a multi-episode sequence, then
\begin{equation}\label{eq:e_co} 
\begin{aligned}
    \arg \min_{\overrightarrow{u}, \overrightarrow{g}, B_0}E_{co} = \arg \min_{\overrightarrow{u}, \overrightarrow{g}, B_0} \norm{B_T^{i-1} - B_0^{i}}_2^2, 
\end{aligned}
\end{equation}
where $T$ is the number of frames in the $i^{th}$ episode. 

\noindent \textbf{Two Persons:} For the case with two persons, while the pose projection constraint remains the same (only this time, applied to multiple poses), the contact term needs to accommodate information about which person the object is in contact with: 
\begin{equation}\label{eq:contact_term} 
    \arg \min_{\overrightarrow{u}, \overrightarrow{g}, B_0} E_c(P) = \arg \min_{\overrightarrow{u}, \overrightarrow{g}, B_0} \sum_{(c, t) \in \mathcal{C}} \norm{P^{c, \delta}_t -  B_t }_2^2. 
\end{equation} 
In this equation, $P^{c,\delta}_t$ indicates the 3D position of the joint $c$ of the person $\delta$ ($\vert\delta\vert$ is the number of people in the scene) at time of contact  $t$. 

\begin{figure}[!th]
    \centering
\includegraphics[width=\linewidth]{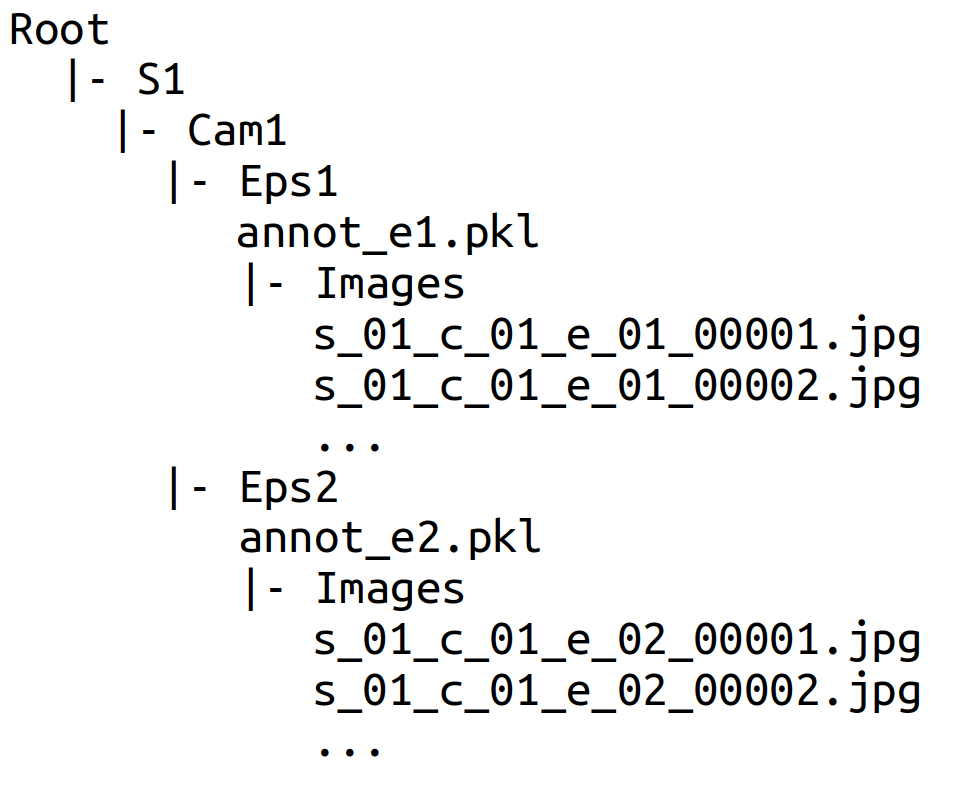}
    \caption{Dataset structure tree}
    \label{fig:structure}
\end{figure}

\section{Dataset Structure} 
Our dataset consists of nine activity sequences (eight single-person and one with two persons) performed by four subjects. 
For each sequence, we have two-three multi-episodes involving one or more episodes in succession. 
We provide annotations and images for up to three camera views. 
The annotations include the 2D and 3D human poses, 2D and 3D object trajectories, the camera calibration parameters and point-of-contact information (frame numbers and joints which are closest to the body at the time of contact). 
The images are of size 1200x877 px.
The structure of the dataset is demonstrated in Fig.~\ref{fig:structure}. 

\end{document}